\documentclass[letterpaper, 10 pt, conference]{ieeeconf}  

\IEEEoverridecommandlockouts                              
\usepackage[numbers]{natbib}
\usepackage{graphicx}
\usepackage{amsmath}
\usepackage{algpseudocode}
\usepackage{algorithm}
\usepackage[algo2e]{algorithm2e}
\usepackage{multirow}
\usepackage[normalem]{ulem}
\usepackage{hyperref}
\usepackage{caption}
\usepackage{subcaption}
\usepackage{lipsum}
\usepackage{soul}

\hypersetup{
    colorlinks=true,
    linkcolor=blue,
    filecolor=magenta,      
    urlcolor=blue,
    citecolor=magenta,
}

\title{\LARGE \bf Context is Everything: Implicit Identification for Dynamics Adaptation}

\author{Ben Evans \\ 
New York University \\
\texttt{benevans@nyu.edu}
\and
Abitha Thankaraj\\
New York University \\
\texttt{abitha@nyu.edu}
\and
Lerrel Pinto \\
New York University \\
\texttt{lerrel@cs.nyu.edu}
}

\begin{document}
\maketitle
\thispagestyle{empty}
\pagestyle{empty}

\begin{abstract}
Understanding environment dynamics is necessary for robots to act safely and optimally in the world. In realistic scenarios, dynamics are non-stationary and the causal variables such as environment parameters cannot necessarily be precisely measured or inferred, even during training. We propose Implicit Identification for Dynamics Adaptation (IIDA), a simple method to allow predictive models to adapt to changing environment dynamics. IIDA assumes no access to the true variations in the world and instead implicitly infers properties of the environment from a small amount of contextual data. We demonstrate IIDA's ability to perform well in unseen environments through a suite of simulated experiments on MuJoCo environments and a real robot dynamic sliding task. In general, IIDA significantly reduces model error and results in higher task performance over commonly used methods. Our code and robot videos are available here \url{https://bennevans.github.io/iida/}

\end{abstract}

\section{Introduction}
Getting robots to adapt to new environments and effectively interact with new objects is a long-standing challenge in robotics. Works in large-scale data collection~\cite{pinto2016supersizing,levine2016learning} and simulation to real learning~\cite{Tobin2017DomainRF,pinto2017asymmetric,peng2018sim,openai2019solving} have offered promise in robot generalization by fitting large parametric models on diverse robotic data. However, such methods are either only able to solve simple manipulation skills such as pushing and grasping, while for more complex skills large training times in the order of a few months are often required~\cite{openai2019solving}.

On the other hand, humans and other biological agents are able to seamlessly generalize to new environments even under adversarial settings~\cite{swaboda2018preschoolers, schulz2012origins}. We hypothesize two reasons for this. First, instead of simple feed-forward processing of inputs, we are able to continuously adapt our belief of the world and reason through past experience with similar environments~\cite{stenning2012human}. Second, our processes for adaptation are versatile and fast. Unlike standard neural network architectures, they can quickly adapt to new experiences and handle varying amounts of information~\cite{smith2005development}. Given that biological systems can generate robust adaptation mechanisms, a natural question arises: How do we endow our robots with such adaptation ability?

Adaptation in robotics has been explicitly studied under the umbrella of online system identification~\cite{Ljung1998,nelles2001nonlinear,Yu-RSS-17}. Concretely, given a dynamic system with unknown physical parameters such as mass, friction, or wind speed, an identification module is optimized to infer these unknowns given observational data from the system. Such inference is empirically fast and robust enough to be deployed on a variety of robotic systems from manipulators to drones. However, a significant assumption made in system identification is that the physical parameters are known and that they can be precisely estimated from observational data.

This assumption breaks down in settings where the number of physical parameters is too large, leaving multiple possible explanations for the observed behavior, or when we cannot directly model all of the environment variations~\cite{zhou2018environment}. 
Moreover, learning-based methods that seek to approximate such identification systems often require measuring the ground truth physical parameters~\cite{Yu-RSS-17, lee2020learning, chebotar2019closing}, which is particularly expensive in settings where we want to adapt to a multitude of environments.

\begin{figure}
  \begin{center}
    \includegraphics[width = \linewidth]{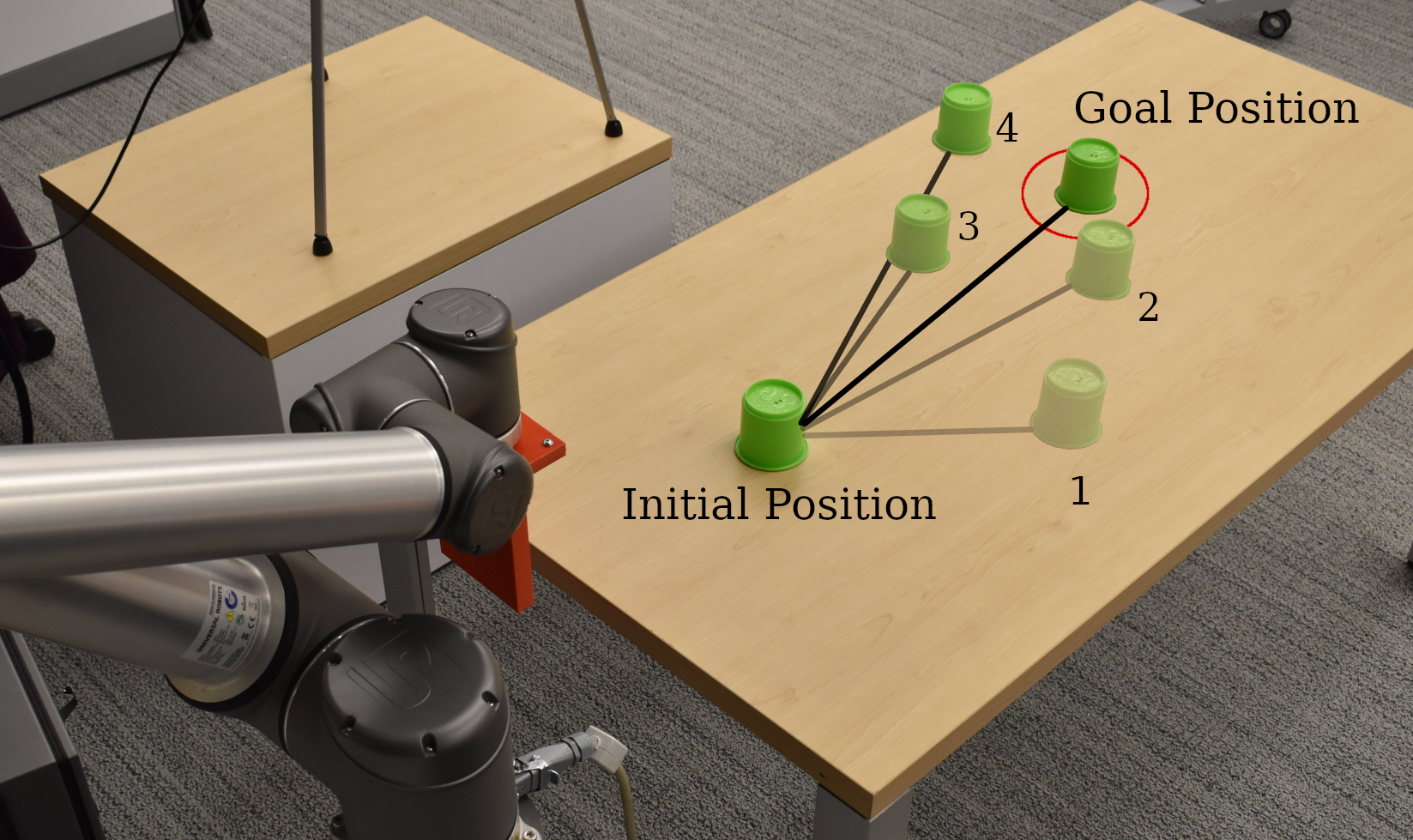}
  \end{center}
  \caption{Adaptation for Robot Sliding. Given an object the robot has not encountered previously, our robot performs implicit identification of the object through IIDA using a handful of interactions on the same object as context points (1-4). This enables dynamic sliding from an initial position within the robot's reach to a user-specified target position outside the robot's reach. }
\label{fig:intro}
\end{figure}

In this work, we present  Implicit Identification for Dynamics Adaptation (IIDA), an approach for adaptation that can identify robotic environments without needing to access the environment's true physical parameters. IIDA consists of two neural networks, one for producing implicit estimates from historical experience, and the second that uses these implicit estimates to predict the dynamics of the environment. 

To train IIDA we first create a context-aware dataset that splits the robotic data across a variety of environments into tuples of transitional experience from the same environment. During training, we provide a variable amount of `context' data from the environment tuple as input to the implicit identification module, while the prediction module gets a single example as input. The prediction loss generated from the prediction module is used to train both modules in an end-to-end fashion such that at test time, we can use new context points to implicitly identify the unseen environment. 

We experimentally evaluate IIDA on a variety of simulated and real-robotic model predictive settings. For single-step prediction on pushing and sliding tasks, we see significant gains over representative baselines in explicit system identification, domain randomization, and meta-learning. Similarly, on multi-step prediction on MuJoCo based OpenAI Gym tasks~\cite{todorov2012mujoco,brockman2016openai} such as Hopper, Swimmer, and Humanoid, we see an average of 63\% reduction in Mean Squared Error on predictions compared to state-of-the-art baselines. The trend holds on our real robot object sliding task (Fig.~\ref{fig:intro}), where a manipulator is tasked with sliding a previously unseen object to a desired goal location on the table. 

To summarize, this paper presents the following contributions. First, we present IIDA, a novel yet simple to implement environment identification technique that does not require any access to ground truth environment parameter information. Second, we show that IIDA can adapt to out-of-distribution environments with just a handful of contextual examples. Third, we run quantitative and qualitative ablation studies to understand the effects of hyperparameters and failure modes. Finally, we demonstrate IIDA on a robotic dynamic manipulation task, where it can slide novel objects across a table within an accuracy of 5cm at 57.1\% success rate, which is higher than the 42.6\% that Domain Randomization achieves. All of our datasets, algorithms, and robot videos will be publicly released on our project webpage.

\section{Related Work}
Our work draws inspiration from a variety of sub-fields in Machine Learning and Robotics such as system identification, adaptation through randomization, and meta-learning. Here, we only discuss the most relevant ones to our work.

\subsection{System Identification}
System Identification (SysID) assumes a parametric analytical model and attempts to find parameters that minimize the difference between the model and the test environment \cite{Ljung1998, nelles2001nonlinear}. The model is then directly used to find controls or to train a policy in simulation. \cite{Yu-RSS-17} first used supervised learning to learn an online system identification model that predicts the environment parameters and then learn a ``universal policy'' conditioned on those parameters to act optimally in the environment. However, for systems with many parameters and without explicit access to their ground truth values during training, obtaining good SysID is challenging~\cite{zhou2018environment}. Instead of explicit SysID, IIDA performs implicit identification and hence does not need to recover the underlying system parameters.

\subsection{Adaptation through Randomization}
Instead of directly modeling the test environment, another common approach, known as Domain Randomization (DR), is to train a single policy on a family of task parameters with the hope of learning a model that can perform well at test time \cite{Tobin2017DomainRF,lowrey_dr,grasping_da,pinto2017asymmetric}. Closest to our work, CaDM \cite{cadm} uses a short history of state-action pairs to produce a latent vector, but can only reason about the most recent timesteps. Policies trained with recurrence or using model ensembles have also been used to manage diversity in environments \cite{openai2019solving, DBLP:journals/corr/RajeswaranGLR16,peng2018sim}. However, such randomization techniques often lead to conservative policies that do not adapt to new environments. More recent work in adaptation~\cite{RoboImitationPeng20,chebotar2019closing, kumar2021rma} overcome this challenge by making online updates to latent environment encodings. This however requires access to the ground truth environment parameters during training. In IIDA we take inspiration from such adaptation techniques and create a framework that does not require access to environment parameters either during evaluation or training.

\subsection{Meta Learning}
Meta-Learning is a framework that explicitly learns how to adapt to meta-tasks given a limited amount of data from the task at test time. Techniques include learning a good initialization for gradient descent such as MAML~\cite{pmlr-v70-finn17a, rajeswaran2019metalearning} and using external memory in the architecture~\cite{pmlr-v48-santoro16}. Most relevant to our adaptation problem, algorithms in the MAML family can use context points from a given new environment and update its model of the world~\cite{finn2017one,kaushik2020fast,rajeswaran2019metalearning, co2021evolving}. In practice, MAML requires second-order optimization and hence needs significantly large amounts of data for effective training. Implicit MAML (iMAML)~\cite{rajeswaran2019metalearning} is empirically more stable, so we use that as a baseline. Other methods learn a neural network prior alongside a simple model and update the simple model using a small amount of data \cite{mbrl_flight, skills_da}. IIDA, on the other hand, does not update the model parameters during adaptation and can hence learn to effectively use a small number of context points from a total of a few hundred training examples.

\section{Background}
\subsection{Formalism for MDP task families}

We consider a family of MDPs that are tuples of the form $M(e) \equiv (\mathcal{S}, \mathcal{A}, \mathcal{T}_e, \mathcal{R})$, where $\mathcal{S}$ and $\mathcal{A}$ are continuous state and action spaces, $\mathcal{R}$ is a reward function, and $\mathcal{T}_e(s' | s, a; e)$ is the transition function. The transition function determines the future state $s'$ given current state $s$, action $a$, and environment factors $e$. The environment factors are distributed according to $e \sim P_E(.)$. This distribution is held fixed during training, while during evaluation the environment factors may be sampled from a different distribution. While $e$ is not directly observable, it affects the environment dynamics. For example, $e$ could be the link lengths and gear ratios of a robot, or the mass and shape of an object in the environment.

\subsection{Dynamics modeling from experience}
Learning dynamics models is a common approach to solving robotic learning problems. Concretely, for a given environment, experience is collected into a dataset $\mathcal{D} = \{(s_i, a_i, s_i')\}_{i=1}^n$ and a parameterized model $f_\theta$ is trained in a supervised fashion to minimize expected reconstruction error 
\begin{equation}
    E_{(s, a, s') \sim \mathcal{D}}[\left\lVert f_\theta(s, a) - s'\right\rVert_2]
    \label{eq:pred}
\end{equation}
In our setting, the same $(s,a)$ can result in vastly different $s'$, depending on the environment parameters $e$. Hence it is crucial for dynamics modelling methods to capture underlying environment information through past experience. 

\begin{figure*}
  \begin{center}
    \includegraphics[width = 0.7\textwidth]{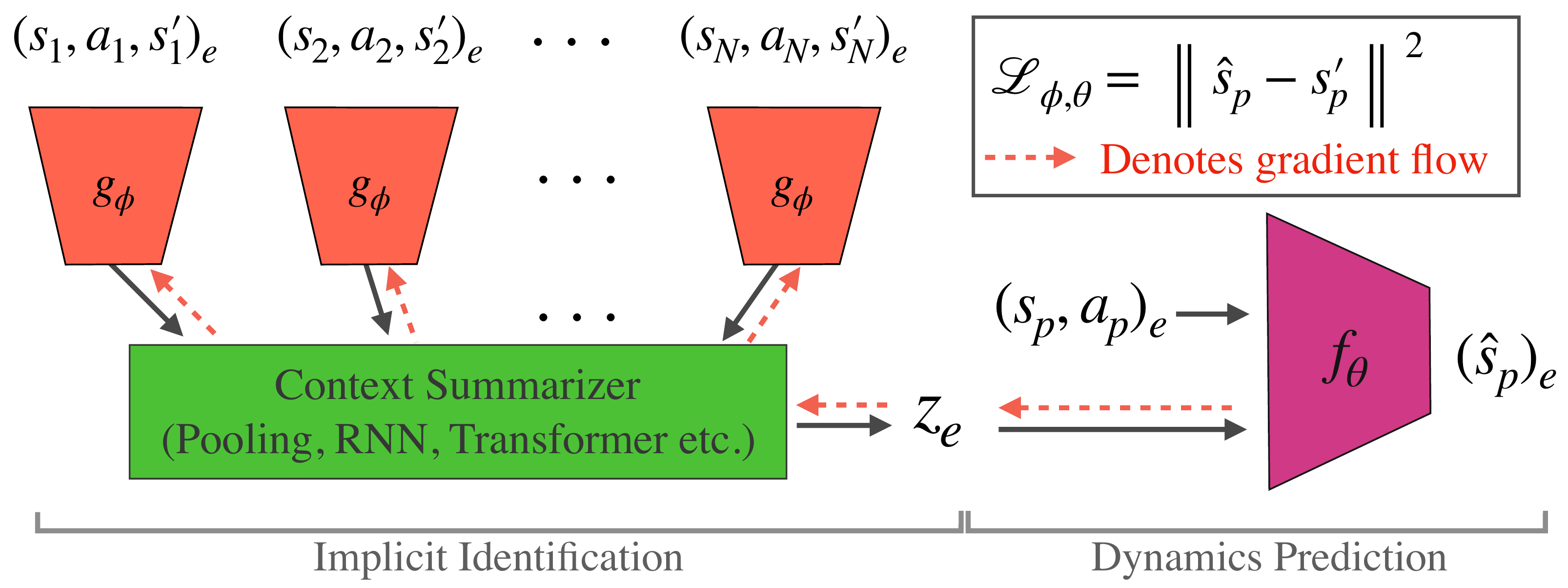}
  \end{center}
  \caption{The Implicit Identification and Dynamics Prediction Models. Context points $(s,a,s')$ from the environment contain identifying information about the dynamics of the world. The Implicit Identification model can take an arbitrary number of context points to implicitly infer properties of the environment, resulting in a context-aware dynamics model.
  }
\label{fig:arch}
\end{figure*}


\section{Implicit Identification for Adaptation}
We would like to learn a model that performs well over the distribution of training environments $e \sim P_E(.)$ without access to the environment parameters $e$. Without access to this ground truth environment information, our model will need to extract equivalent information from past experience, which we call `context'. Such a contextual learning setup affords a collection of datasets $D=\{\mathcal{D}_e\}$ from unknown parameters $e$. Note that this mimics robotic problems where we know that the environment has changed, but do not know the exact factors that have changed. For instance, in object manipulation, when the robot interacts with a new object, we know that the properties of the environment have changed; but we do not know the exact factors such as mass, friction, roughness, or size of this object.

Many standard formulations capture and attempt to solve this problem. In the POMDP setting~\cite{braziunas2003pomdp}, where recurrent neural networks are commonly used, we could augment the state to include environment parameters $e$ that are inferred from observations. Another common approach used is Meta Learning~\cite{pmlr-v70-finn17a}, where models can be updated using a small amount of test data. In this work, we present an approach that combines key ideas from these broad lines of work. Similar to meta-learning, we adapt the robot's prediction model. However, we do this without updating the parameters of the model. Instead, similar to POMDP settings, we infer the environment information by summarizing the context examples for that given environment. Since the environment information we obtain is not the true factors of the environments, we call this process Implicit Identification. Once the environment has been identified, we use the implicit factor $z_e$ to perform downstream dynamics modeling. The full pipeline is illustrated in Fig.~\ref{fig:arch}.  In the following sections, we describe the two parts of our model, Implicit Identification and Dynamics Prediction along with the training and inference procedure in detail.

\subsection{Context Modeling and Summarization}
Since we have no knowledge of the true environment parameters, we must infer information about the environment through the interaction data. Tuples of $(s,a,s')$, which we call context points, contain information about the dynamics of the environment. Our method selects $N$ context points $\{(s_i, a_i, s_i')\}_{i=1}^N$ at random from the current environment, passes each of them through a context encoder $g_\phi$ to produce $N$ latent vectors $\{z_i\}_{i=1}^N$, which are combined in an order-invariant fashion to produce a single latent $z$. Our model works for arbitrary values of $N$, improving performance as we get more context. The latent factor $z_e$ contains identifying information about the environment without explicitly modeling any of the environment parameters, and hence implicitly reasoning about the dynamics of the world. 

As noted in Fig.~\ref{fig:arch} there are several techniques we can use to summarize order-invariant context points. In this work we investigate three such methods:

\paragraph{Average Pooling} Our simplest context encoder is a simple feed-forward neural network with average pooling. We pass the context points through the same encoder $g_\phi$ and average the resulting vector to produce latent $z_e$. It is easy to implement and requires minimal computation while providing competitive performance. While powerful, this model treats each context point equally, which may be undesirable if some of the context points contain little information about the environment e.g. the start and end state of a sliding maneuver is exactly the same.

\paragraph{Recurrent Neural Networks (RNNs)} To improve upon Average Pooling, we investigate the use of RNNs. Typically, RNNs are used to model sequential data in the form $(s_t, a_t), (s_{t+1}, a_{t+1}), \ldots, (s_{t+H}, a_{t+H})$, leveraging the ordering of the data to improve performance. In a single-step environment, however, there is no way to exploit correlations between adjacent inputs.
We take a different approach, treating randomly ordered context tuples $\{(s_i, a_i, s_i')\}_{i=1}^N$ as a sequence. We use a linear projection of the last hidden state as our latent $z_e$. Instead of learning temporal correlations, the RNN learns to extract useful information from each context point in the sequence. The model then has the capacity to decide what information to extract from each observation, conditioned on the other context points it has seen. In practice we use an LSTM~\cite{LSTM} for its stable training property.

\paragraph{Transformers} Transformers \cite{NIPS2017_3f5ee243} are an attention-based model often used for sequence modelling. We omit the positional encoding used for sequential tasks giving us an order-invariant processing module. Because each input is allowed to attend to all the other inputs, the model is able to weight context points based on what it thinks is important, allowing it to discard uninformative contexts and non-linearly combine information from informative ones.

\subsection{Predictive Modeling with Context}
To investigate the effectiveness of IIDA and solve challenging robotic problems we learn predictive models on observational data. In contrast to standard predictive models $f_\theta(s,a)$, we include the latent $f_\theta(s,a;z_e)$ so it has information about the current environment. Once such a predictive model is learned, we can perform standard model-based optimization to train robotic behaviors to maximize rewards. Since this work focuses on Implicit Identification, we limit our behavior optimization to simple Cross-Entropy Maximization~\cite{shore1980axiomatic}. 

\subsection{Training IIDA}
Given the Implicit Identification and Dynamics Prediction modules, we train the context summarizer $g_\phi$ and prediction model $f_\theta$ with end-to-end gradient descent.
We use the standard reconstruction loss:
\begin{equation}
   E_{\mathcal{D}_e \sim D, \{(s_i, a_i, s_i')\}_{i=0}^N \sim \mathcal{D}_e}[\left\lVert f_\theta(s_0, a_0, z_e) - s_0'\right\rVert ]  \\
    \label{eq:IIDA}
\end{equation}
where $z_e = g_\phi(\{(s_i, a_i, s_i')\}_{i=1}^N)$.
The identification model $g_\phi$ is hence encouraged to encode useful information in the latent that results in a model that can vary its predictions given interaction data from the environment. Note that unlike contemporary adaptation techniques~\cite{RoboImitationPeng20,chebotar2019closing}, no additional losses are needed to train $g_\phi$, which makes IIDA straightforward to implement and optimize.

\subsection{Inference with IIDA}
After training, IIDA can quickly adapt to unseen variations in the environment. Without any context, we pass in an arbitrary `no-context' vector for $z_e$. We can iteratively collect more context points and include them in our model to improve predictions. Each interaction may only contain partial information, so the ability to utilize multiple context points allows us to better capture information about the environment. Since IIDA does not make assumptions on the temporality of the context points, it is applicable to both temporally related context examples or independently sampled ones, without significant implementation changes. 

\section{Experiments}
\begin{figure}
  \begin{center}
    \includegraphics[width = \linewidth]{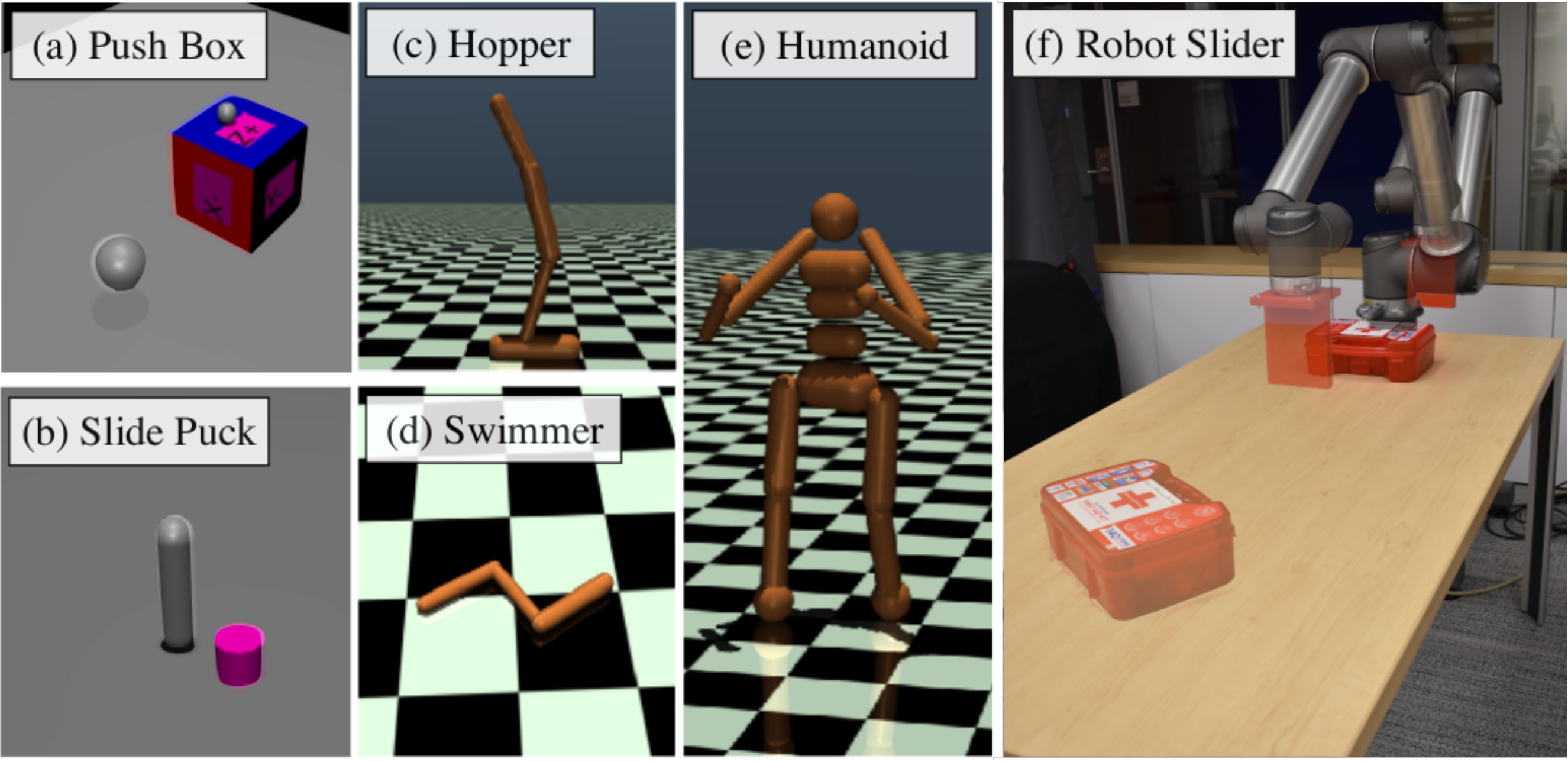}
  \end{center}
  \caption{ (a) The Push Box environment. Along with varying frictions and masses, the mass at the top box is offset by a constant amount, resulting in skewed pushes. (b) The Slide Puck environment. (c-e) A sample environment for each of the MuJoCo environments. (f) Our robotic experiment setup.}
\label{fig:envs}
\end{figure}

In the previous sections, we have described IIDA, a simple adaptation technique that performs implicit identification using contextual examples. In this section, we seek to experimentally investigate the following questions: Can IIDA adapt to different environments? How does IIDA's performance vary based on context? How well does IIDA adapt in real-robotic settings? Is IIDA actually identifying anything? We will answer these questions through experimental evaluations on both simulated and real-robot settings.

\subsection{Experimental Setup}
We consider learning a model on a number of environments simulated using the MuJoCo simulator~\cite{todorov2012mujoco} including 1-step prediction tasks and multi-step environments from OpenAI Gym~\cite{brockman2016openai}. We also verify our method with a real robot experiment.
Details of the three broad environment setups are described below and additional details can be found in Section \ref{sec:ap_details} of the Appendix.

\begin{table*}[t]
\caption{We report the MSE $\pm$ std on prediction over 3 seeds on the test set for the baselines and IIDA. }
    \centering
\begin{tabular}{c|c|c|c|c|c|c}
\multirow{2}{*}{Environment (Scale)} & \multirow{2}{*}{DR} & \multirow{2}{*}{iMAML} & \multirow{2}{*}{Explicit ID} & \multicolumn{3}{c}{IIDA}                             \\ \cline{5-7} 
                                     &                     &                        &                              & Average         & RNN              & Transformer     \\ \hline
Slidepuck ($10^{-4}$)                & 18.64 $\pm$ 4.92    & 17.68 $\pm$ 7.41       & 3.90 $\pm$ 0.79              & 4.42 $\pm$ 0.80 & \textbf{2.78 $\pm$ 0.55}  & 8.43 $\pm$ 4.22 \\
Pushbox ($10^{-3}$)                  & 11.79 $\pm$ 4.24    & 24.19 $\pm$ 0.11       & \textbf{4.95 $\pm$ 1.39}              & 6.73 $\pm$ 3.32 & 7.17 $\pm$ 0.68  & 9.56 $\pm$ 0.92 \\
Hopper ($10^{-3}$)                   & 5.53 $\pm$ 0.57     & 92.15 $\pm$ 16.88      & 5.87 $\pm$ 0.65              & 4.62 $\pm$ 0.73 & \textbf{4.13 $\pm$ 0.59}  & 8.33 $\pm$ 4.23 \\
Swimmer ($10^{-3}$)                  & 45.47 $\pm$ 6.15    & 28.12 $\pm$ 8.20       & 8.91 $\pm$ 2.47              & \textbf{1.72 $\pm$ 0.49} & 10.23 $\pm$ 6.64 & 3.21 $\pm$ 2.47 \\
Humanoid ($10^{-1}$)                 & 8.57 $\pm$ 2.50     & 11.38 $\pm$ 4.03       & 7.28 $\pm$ 3.17              & 2.39 $\pm$ 0.38 & 3.04 $\pm$ 0.92  & \textbf{2.34 $\pm$ 0.44} \\
Robot ($10^{-3}$)                    & 6.30 $\pm$ 0.38     & 7.21 $\pm$ 2.24        & N/A                          & \textbf{3.44 $\pm$ 0.20} & 3.50 $\pm$ 0.25  & 3.81 $\pm$ 0.08
\end{tabular}
    \label{tab:mses}
\end{table*}

\textbf{Single-step Environments.} We create two environments for learning a single-step prediction model: Push Box and Slide Puck. 
\begin{enumerate}
    \item \textbf{Push Box:} We initialize a box in a random $(x,y)$ location and apply a push to the box with a sphere for a fixed distance. The action space consists of the $(x,y)$ location of the start of the push, the angle, and the push velocity. We take the initial state and action and run for 2000 timesteps. The box has a point mass on top of it that when offset, changes the friction normal and how the push behaves. We vary the point mass's COM-offset, box mass, and friction, resulting in a diverse set of behaviors. Because the angle, $\theta$, of the box can wrap around, we use an observation function that contains $(x,y, \sin(\theta), \cos(\theta))$ for the box. We collect data from 100 different environments with 2000 actions in each for training and generate 20 novel validation and test environments, each with 100 actions for testing.
    \item \textbf{Slide Puck:} This environment has a similar action space, but unlike in Push Box, the object is not in contact with the pusher for the duration of the push and instead slides across the table. We vary puck mass and friction, as well as wind, table tilt, and the simulated damping coefficient. We take the initial state and action and allow the simulation to run until the push is complete and the object stops. The resulting end state is used as the label for training. We collect from 1000 environments, but only collect 10 actions in each, generating 100 validation and test environments with 10 actions each to evaluate the generalization ability of our model.
\end{enumerate} 

\textbf{Multi-step Environments.}
We modify the standard OpenAI gym~\cite{brockman2016openai} tasks Hopper, Swimmer, and Humanoid to include variations in the dynamics. Fig.~\ref{fig:envs} shows one such variation in each environment. To collect data, we train a policy in the original environment using Soft Actor-Critic~\cite{DBLP:journals/corr/abs-1801-01290} and roll it out for the entire length of the environment. We then perform a relabeling procedure for each $(s,a)$, setting the environment factors and simulating one step forward to produce a new $s'$.
\begin{enumerate}
    \item \textbf{Hopper:} We generate 100 paths and split them into 80 train, 10 validation, and 10 test paths. We generate 10 environments for each set of paths, varying the mass and length of the links.
    \item \textbf{Swimmer:} Similar to Hopper, we generate 100 paths and split them into 80 train, 10 validation, and 10 test paths. We generate 10 environments for each set of paths, varying the mass and length of the links.
    \item \textbf{Humanoid:} We generate 50 paths and split them into 40 train, 5 validation, and 5 test paths. We generate 25 environments for each set of paths, symmetrically varying the mass and length of the arms, and the horizontal offset of the legs and feet.
\end{enumerate}
To add to the complexity of the problem, we do not sample environment parameters uniformly over the range of possible parameters. Instead, we uniformly divide each varying parameter into 30 ribbons of uniform size and divide the values randomly between the train, validation, and test environments. We give 80\% of the ribbons to the train environment and split the remaining ribbons between validation and test. This means the environment parameters we test on are completely unseen to the model during training, so the model must learn to interpolate and extrapolate between the different environments in order to predict well. We perform the entire dataset generation and relabeling procedures on 3 independent seeds to demonstrate consistency across different instances of the data split.

\textbf{Robot Sliding: }To test our model in the physical world, we design a robotic experiment using a UR10 robot modeled after the Slide Puck environment. The robot moves its slider to a start position determined by the angle at which the object is to be pushed and selects the target tool velocity. The robot translates its slider forward at that velocity by a fixed distance, allowing the object to slide to various $(x,y)$ coordinates on the table. 
We collect a dataset of 20 slides for 27 train, 5 validation, and 10 test objects with varying properties, selecting actions uniformly at random, and save the start and end $(x,y)$ positions. The start and end positions are detected using blob detection on the start frame and end frame of the video recorded. The train set includes slides on both a tablecloth and the table, while the test set includes slides using previously unseen objects on the table, the table cloth, and on a foam surface not seen during training. 

\subsection{Baselines}
We compare against the following baselines:
\begin{enumerate}
    \item \textbf{Domain Randomization (DR)} is a standard technique in which a single neural network is trained in across all environments in the train set to be robust to environment changes \cite{Tobin2017DomainRF}\cite{peng2018sim}.
    \item \textbf{iMAML} is an algorithm in the MAML family, a common approach to Meta-Learning that we compare against~\cite{rajeswaran2019metalearning}. We allow it to update its parameters using the same number of context points as our models. We tried training vanilla MAML~\cite{pmlr-v70-finn17a}, but were unable to get it to converge for our problems.
    \item \textbf{Explicit Identification} is a ground truth model that has access to the true underlying environment parameters at both train and test time~\cite{Yu-RSS-17,RoboImitationPeng20,Kumar2021}.
\end{enumerate}
We also implemented a PEARL-style probabilistic encoder~\cite{pmlr-v97-rakelly19a}, but it failed to produce competitive results, hence we omit it for conciseness.

\begin{figure*}
  \begin{center}
    \includegraphics[width = 0.9\textwidth]{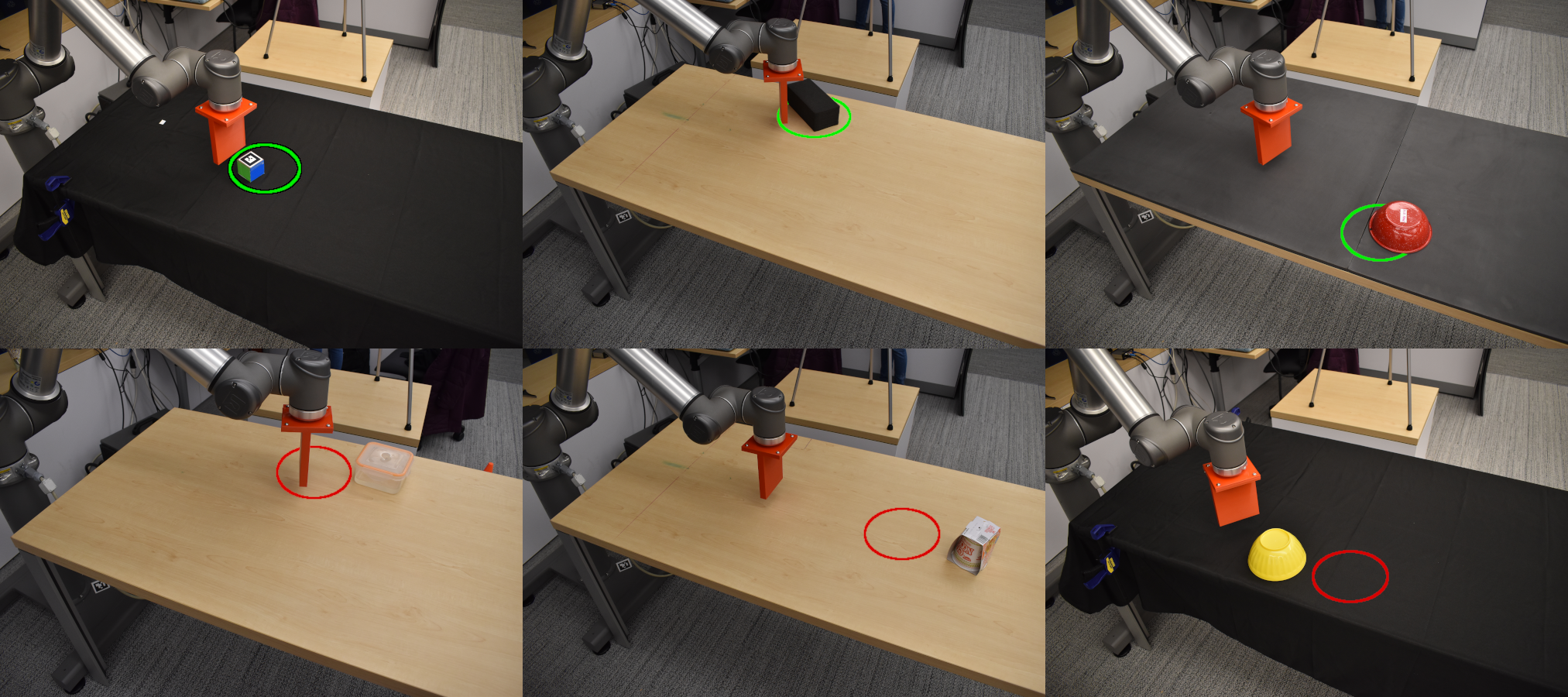}
  \end{center}
  \caption{We illustrate the results of running IIDA on our robot sliding task. The circle represents the target goal location of sliding. Each object is presented with four context points without goal information. In most cases, the objects reach the target location. However, for some objects like the Cup Noodle, Tupperware 8oz, and Cloth Yellow Bowl (Bottom), it misses the goal location by a small margin.
  }
\label{fig:robot}
\end{figure*}

\subsection{Training Details and Network Architectures}
All our dynamics prediction models are two-layer feedforward networks with hidden sizes of 256 and are optimized using Adam~\cite{Kingma2015AdamAM} with a learning rate of $10^{-3}$ and a batch size of 64. For the IIDA models, we append the output of the encoder to the input of the dynamics models. We describe the 3 encoders we experimented with below:
\begin{enumerate}
    \item \textbf{Average Pooling} is a one-layer network with width 256. We concatenate each individual context into one vector, pass it through the network, and simply average the output to produce the latent.
    \item \textbf{RNN} is an LSTM with two layers of hidden size 256. We pass the context in as if it were a sequence and take the last hidden state as the context vector.
    \item \textbf{Transformer} uses a single layer of multi-headed attention of size 120 with 5 heads. We project the output to the required dimension using a linear layer.
\end{enumerate}
We pick a latent dimension of 8 for all IIDA models. Simulated single-step environments are trained with 8 context points, multi-step environments with 128, and robot experiments with 4. We implement our models in PyTorch~\cite{NEURIPS2019_9015} and will release all code and environments. We modify an existing PyTorch implementation of iMAML~\cite{rajeswaran2019metalearning} to work for regression. We use two layers of size 512, outer and inner LRs of $2.5 \times 10^{-3}$ and $1\times 10^{-3}$, and 16 update steps.

\subsection{Can IIDA adapt to different environments?}
To quantify IIDA's ability to adapt to new environments, we consider learning predictive models on a number of training parameters and testing them on new data with unseen variations. We test on the model with the best validation loss, which like the test set, contains new data and environment parameters. All results can be found in Table \ref{tab:mses} and are discussed below. 

\subsubsection{Single-step Environments}
In the Slide Puck environment, all IIDA models outperform the DR and iMAML baselines, with the RNN model outperforming the Explicit Identification model with access to ground truth environment parameters. The best IIDA model reduces error by 84\% compared to the best baseline without access to the explicit world parameters.
In the Push Box environment, the IIDA models also outperform the baselines, reducing error by 43\%. The IIDA models fail to beat the Explicit ID model, which we expect since only 5 factors are varying and one, the center of mass offset, drives a majority of the behavior. Having that number would be highly informative when compared to a few context pushes.

\subsubsection{Multi-step Environments}
IIDA outperforms all baselines in all environments, reducing error from better of DR and iMAML by 25\%, 93\%, and 72\% for the Hopper, Swimmer, and Humanoid environments, respectively. We even outperform the Explicit Identification model. By reasoning about the behavioral properties of the environment instead of the true underlying parameters, we generalize to a wider range of variations. While initially counterintuitive, this highlights that challenge that even explicit identification faces when the environment is out of distribution, or when there are many factors that can change. Although the environment can take on many parameters, the possible behavior of the system may lie on a lower-dimensional manifold.

The performance gap between DR and IIDA is much smaller for Hopper than the other multi-step environments. We believe that this is because the dynamics of the Hopper cannot vary as much as in the Swimmer and Humanoid environments. A change of mass or length in the Swimmer environment will affect the inertia of the entire body in the unconstrained 2D space more than the Hopper which is on the ground for a portion of time and is affected by gravity.
\subsubsection{Robot Environment}
We report the average Euclidean error between model predictions and end locations in the static dataset. We do not know the true parameters of the objects, so we omit the Explicit ID model. On the held-out pushes for new objects, the DR model achieves an average error of 7.9cm, and the iMAML model achieves 8.5cm of error. The IIDA models are able to all beat the baselines, with average errors of 5.8, 5.9, and 6.2 cm for the average, RNN, and transformer models, respectively, corresponding to a 26\% decrease in error. We believe the limited capacity of the average IIDA model actually helps the model generalize since we have significantly fewer training points compared to our simulated experiments.

We believe that iMAML especially struggles in this setting due to the low amount of context. It is only given 4 context observations to update its parameters, which may not be enough data to modulate the model's parameters to generalize to new objects.

\begin{table}[]
\caption{Sliding success rate for the test objects. We attempt to reach 20 goal positions, fixed for each object so the models have a fair comparison. The test set contains slides on tabletop-material combinations not present in training.}
    \centering

\begin{tabular}{c|c|c|c|c|c}
\multirow{2}{*}{Object}      & \multirow{2}{*}{DR} & \multirow{2}{*}{iMAML} & \multicolumn{3}{c}{IIDA} \\ \cline{4-6} 
                             &                     &                        & Avg  & RNN  & Tfm \\ \hline
sponge                       & 50.0                & 60.0                   & 50.0 & 50.0 & 65.0        \\
cup\_o\_noodles              & 15.0                & 15.0                   & 25.0 & 40.0 & 30.0        \\
cloth\_hand\_cube            & 60.0                & 15.0                   & 90.0 & 90.0 & 95.0        \\
cloth\_plastic\_bowl\_yellow & 35.0                & 20.0                   & 50.0 & 25.0 & 45.0        \\
tupperware\_8oz              & 28.6                & 21.4                   & 35.7 & 42.9 & 64.3        \\
timer\_in\_box               & 25.0                & 10.0                   & 75.0 & 55.0 & 65.0        \\
red\_bottle\_square\_up      & 15.0                & 5.0                    & 40.0 & 25.0 & 25.0        \\
first\_aid                   & 70.0                & 15.0                   & 50.0 & 55.0 & 50.0        \\
foam\_red\_bowl              & 50.0                & 10.0                   & 45.0 & 75.0 & 55.0        \\
green\_cup                   & 65.0                & 30.0                   & 60.0 & 65.0 & 60.0        \\ \hline
Overall success rate         & 42.6                & 20.8                   & 53.7 & 53.9 & 57.1       
\end{tabular}
    \label{tab:success}
\end{table}

\subsection{How well does IIDA adapt in real-world settings? }
We test IIDA in a real world robotic experiment. We take the best validation model learned from the object datasest and use it to reach goal positions on the table. For each test object, we use the end positions from the already collected context points as fixed the goals for all models, ensuring we know the goals are reachable and are consistent across all models.
We get actions from our model by minimizing the error between the target and predicted next state over the action space $a = \underset{a}{\arg\min} ||f_\theta(s, a, z_e) - s'||$ via the cross-entropy method. We test the model by executing the found action, defining success as sliding the object within a 5cm radius of the goal. Details can be found in Section \ref{sec:robot_exp_det} of the Appendix.

We consider a slide a success if the object's end position is within 5cm of the goal. Success rates for different models can be found in Table \ref{tab:success} and qualitative results in Fig.~\ref{fig:robot}.
The IIDA models have a higher success rate, 53.7\%, 53.8\%, and 57.1\% for the average, RNN, and transformer encoders, respectively, vs 20.8\% and  42.6\% for iMAML and the model without any context respectively. Although the transformer model has the highest MSE among the IIDA models, it performs the best in the real world when used to perform the slides. Although lower MSE generally correlates with higher success in the real world, we conjecture that the transformer architectures are more amenable to searching for actions that are likely to hit the target \cite{janner2021sequence}.

\begin{figure}[t]
      \begin{center}
        \includegraphics[width = 0.9\linewidth]{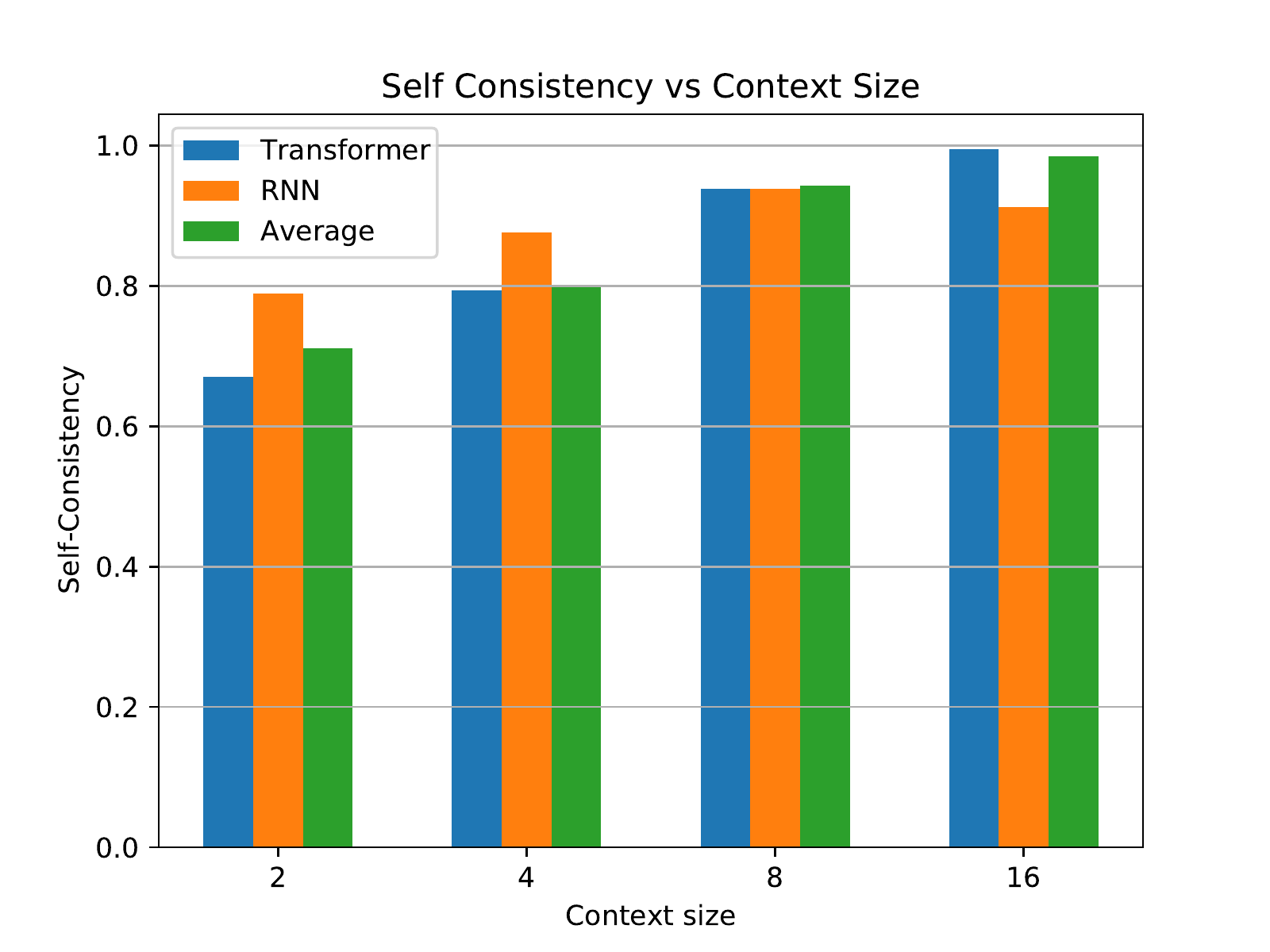}
      \end{center}
    \caption{Self consistency vs context size in the latent space for each of the IIDA models on the robot sliding task. The RNN is more consistent at lower context sizes, but fails generalize to higher context sizes. The average and transformer models are monotonically consistent with more data.}
    \label{fig:nearest_neighbors}
\end{figure}

\subsection{How difficult is the robot sliding adaptation task?}
To put the robot experiment into more context, we ask three volunteers to compete against the robot in a sliding task. We ask the humans to slide 3 objects of varying difficulty (green\_cup, tupperware\_8oz, and cup\_o\_noodles) to 5 different locations on the table. We allow them to play with the object for 30 seconds to provide them context and then ask them to slide to the goal locations. The IIDA transformer model is able to achieve an average error of 8.5 cm, 10 cm, and 14 cm for green\_cup, tupperware\_8oz, and cup\_o\_noodles, respectively, while the humans get an average error of 19 cm, 69 cm, and 38 cm. This highlights the difficulty of our adaptation task with small amounts of contextual information.

\begin{figure*}[t]
    \centering
    \includegraphics[width=0.8 \textwidth]{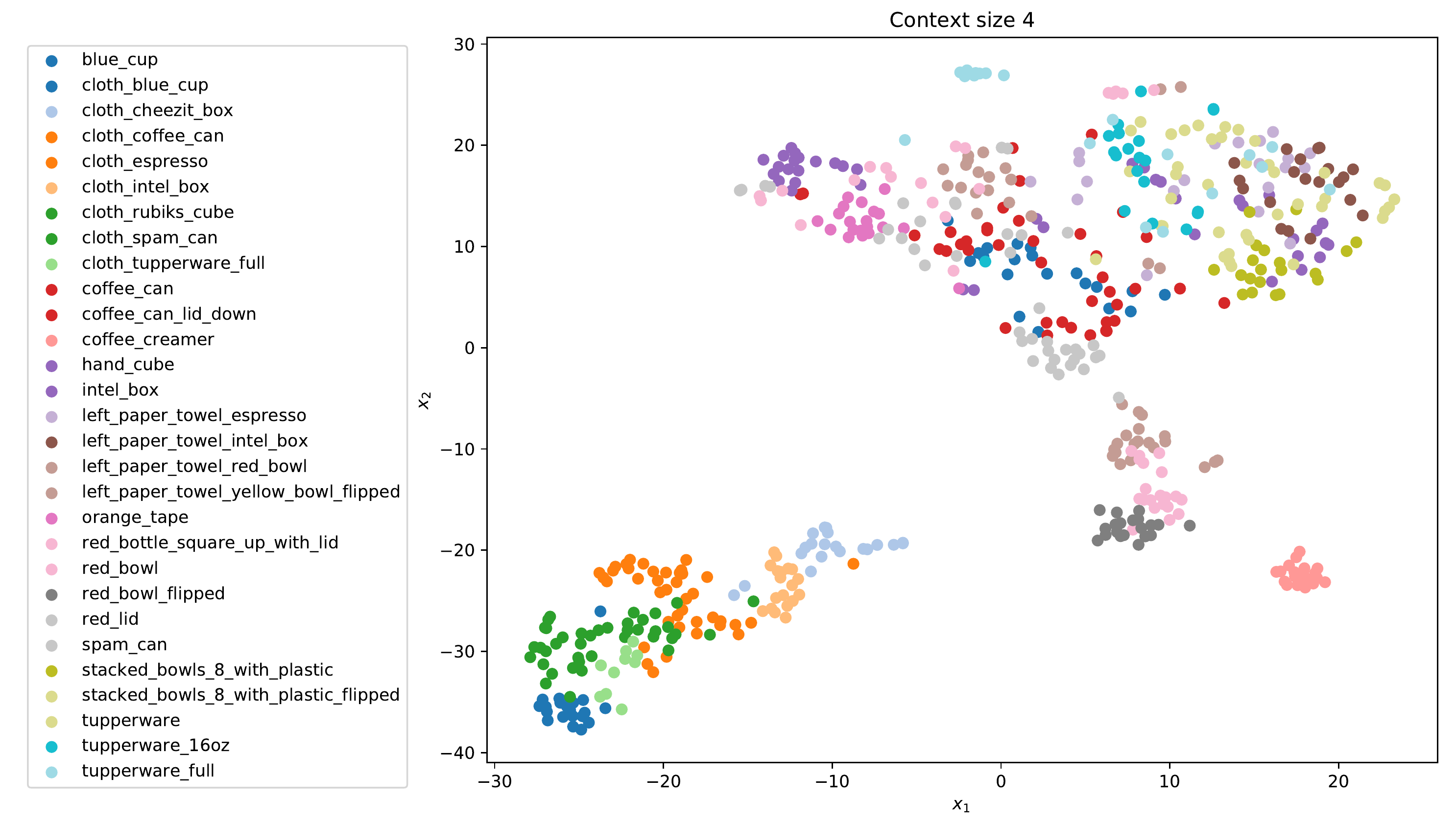}
    \caption{Visualization of the training objects' latent space for the transformer model with the best validation error. In addition to similar latents for the same objects, we can see distinct clusters corresponding to objects with distinct dynamics.}
    \label{fig:latent_tsne_tfm_train}
\end{figure*}
\subsection{How does IIDA's performance vary with size of context?}
For our experiments, we chose a fixed number of context points, however, an advantage of our model is it can vary how much context it receives. For instance, given the Average model that was trained on the robotic sliding task with 4 context points, we give it a varying number of points from $[2,4,8,16]$ for testing. We get test errors of $[4.5, 4.4, 3.9, 3.8] \times 10^{-3}$, a monotonic drop in error with an increasing number of context points. For the RNN, we get errors of $[6.3, 5.4, 4.7, 4.7] \times 10^{-3}$, while the transformer gets errors of $[4.0, 3.6, 3.3, 3.4] \times 10^{-3}$. This highlights the ability of the transformer to generalize to and effectively use more context points. The RNN is not able to generalize since it treats the context as a sequence and has not learned to encode sequences of varying lengths.

\subsection{Does IIDA actually implicitly identify environments?}
IIDA performs well numerically, but it is not immediately clear if the context encoder is indeed 'identifying' the environment. To investigate this, we perform experiments analyzing the learned latent spaces. 

\subsubsection{Latent Nearest Neighbors}
\label{sec:nn}
To test if the learned latent space produces consistent values in the same environment, we create a dataset of latents from the robot sliding test set. For each object, we subsample the context points 20 times to produce 20 latents per object. Then, for each latent, we find the closest latent in the dataset and consider it a success if the object is the same for both latents. We call the success rate self-consistency and plot the metric versus the number of context points for each model in Fig. \ref{fig:nearest_neighbors}. Note that each model is only trained for 4 context points. Even with 2 context points, we have relatively high self-consistency, much better than picking a latent at random (0.1), and adding more context makes it more self-consistent. The RNN model is more consistent with less data but does not generalize to higher numbers of context points, while the average and transformer models are initially less consistent, but are able to generalize given more data.

\subsubsection{Visualization of the Latent Space}
\label{sec:vis_latent}
We investigate if the latent space learned for objects is interpretable by reducing the dimension using t-SNE \cite{tsne} and plotting it in two dimensions. We use the same dataset as in Section \ref{sec:nn} and reduce the dimension, plotting the latents from the same object with the same color. As we can see in Fig. \ref{fig:latent_tsne_tfm_train}, for the transformer encoder the object latents are close to one another, confirming the results of our nearest neighbor experiment and indicating we are learning some underlying properties about the object's dynamics. We can see distinct clusters corresponding to different sliding dynamics. Notably, the objects slid on cloth are all clustered close to one another, showing that objects with similar sliding properties are similar in latent space. Visualizations of additional models can be found in Section \ref{sec:ap_latent} of the Appendix.

\section{Limitations, Discussion and Future Work}
In this paper, we have presented a simple technique for contextual adaptation and demonstrated its performance on a variety of simulated and real-robot tasks. We believe it opens up several avenues for future research, particularly in areas where our method is insufficient. For instance on a few objects such as `red\_bottle\_square\_up' all IIDA context summarizers perform poorly. One hypothesis for this is that this object is quite far away from the training set of objects and hence would require more context points. In general, the relative performance gap was larger in simulation than in the robot experiment. We believe that this is due to the sheer amount of data available and that collecting more training data would improve the relative performance of IIDA.

Although we have shown that IIDA produces strong results for adaptation in prediction-based tasks, our framework is general and can be applied to a variety of other domains:
\begin{enumerate}
    \item As presented, IIDA learns dynamics models that are conditioned on randomly collected context points. Although the transformer's attention mechanism allows it to weigh different context points differently, allowing the model to actively select actions to collect its own context points, rather than sampling from a fixed set could improve performance.
    \item To get actions out of our dynamics models, we use an optimization procedure. Instead, we could learn policies directly, using policy gradients instead of minimizing MSE, for example, to potentially provide additional performance gains. This would be especially important in environments where we want to be able to act, but modeling the system directly is infeasible.
    \item Complementary to both the model and policy learning is the idea of using IIDA in a simulation to real fashion. Rather than hoping that a DR model generalizes from simulation, you could treat the real world as an environment of its own, collecting context, or even including it as part of the training procedure.
\end{enumerate}

\section{Acknowledgements}
We thank Kendall Lowrey, Denis Yarats, David Brandfonbrener, and Anthony GX Chen for their feedback on early versions of the paper. This work was supported by grants from Honda, Amazon, and ONR award numbers N00014-21-1-2404 and N00014-21-1-2758.



\bibliographystyle{IEEEtran}
\bibliography{IEEEabrv,references}

\newpage
\section{Appendix}
\subsection{Additional Environment Details}
\label{sec:ap_details}
Here we provide the parameters we vary and the ranges for each environment.
\begin{table}[H]
    \centering
    \begin{tabular}{c|c|c}
        Parameter & Low & High \\
        \hline
        puck\_mass & 0.15 & 0.4 \\
        floor\_friction & 0.02 & 0.1 \\
        puck\_friction & 0.02 & 0.1 \\
        wind\_x & -5.0 & 5.0 \\
        wind\_y & -5.0 & 5.0 \\
        table\_tilt\_x & -1.0 & 1.0 \\
        table\_tilt\_y & -1.0 & 1.0 \\
        damping & 0.0001 & 0.075 \\
        
    \end{tabular}
    \caption{Varied parameters for the Slide Puck environment}
    \label{tab:env_push_box}
\end{table}
\begin{table}[H]
    \centering
    \begin{tabular}{c|c|c}
        Parameter & Low & High \\
        \hline
        com\_offset & -0.14 & 0.14 \\
        box\_mass & 0.5 & 2.5 \\
        box\_friction & 0.7 & 1.3 \\
        floor\_friction & 0.7 & 1.3 \\
        pusher\_friction & 0.7 & 1.3 \\
    \end{tabular}
    \caption{Varied parameters for the Push Box environment}
    \label{tab:env_push_box}
\end{table}
\begin{table}[H]
    \centering
    \begin{tabular}{c|c|c}
        Parameter & Low & High \\
        \hline
        torso\_mass & 2.0 & 4.5 \\
        thigh\_mass & 2.5 & 4.5 \\
        leg\_mass & 1.5 & 3.5 \\
        foot\_mass & 4.0 & 6.0 \\
        torso\_length & 0.04 & 0.06 \\
        thigh\_length & 0.04 & 0.05 \\
        leg\_length & 0.03 & 0.05 \\
    \end{tabular}
    \caption{Varied parameters for the Hopper environment}
    \label{tab:env_push_box}
\end{table}
\begin{table}[H]
    \centering
    \begin{tabular}{c|c|c}
        Parameter & Low & High \\
        \hline
        torso\_mass & 25.0 & 45.0 \\
        mid\_mass & 25.0 & 45.0 \\
        back\_mass & 25.0 & 45.0 \\
        torso\_length & 0.08 & 0.12 \\
        mid\_length & 0.08 & 0.12 \\
        back\_length & 0.08 & 0.12 \\
    \end{tabular}
    \caption{Varied parameters for the Swimmer environment}
    \label{tab:env_push_box}
\end{table}
\begin{table}[H]
    \centering
    \begin{tabular}{c|c|c}
        Parameter & Low & High \\
        \hline
        thigh\_offset & 0.05 & 0.2 \\
        foot\_offset & -0.5 & -0.4 \\
        upper\_arm\_length & 0.1 & 0.28 \\
        lower\_arm\_length & 0.1 & 0.28 \\
        upper\_arm\_mass & 0.5 & 3.0 \\
        lower\_arm\_mass & 0.75 & 4.0 \\
    \end{tabular}
    \caption{Varied parameters for the Humanoid environment}
    \label{tab:env_push_box}
\end{table}

\subsection{Robot Experiment Details}
\label{sec:robot_exp_det}
\textbf{Dataset:} We collect 20 pushes per object for all objects in the dataset. (With the exception of the object labelled as tupperware\_8oz, for which we collected 13 pushes.) To prevent objects from sliding off the table during experiments, we initially determine the minimum and maximum feasible velocity per object and use these to linearly interpolate the maximum possible velocity for the pushing angle.

\textbf{Sliding experiment MSE:} Here we report the average Euclidian distance between goal end state and actual end state for each of the models and objects. This includes errors from both finding the actions with CEM and noise in the sliding process itself.
\begin{table}[H]
\begin{tabular}{c|c|c|c|c|c}
\multirow{2}{*}{Object}      & \multirow{2}{*}{DR} & \multirow{2}{*}{iMAML} & \multicolumn{3}{c}{IIDA} \\ \cline{4-6} 
                             &                     &                        & Avg  & RNN  & Tfm \\ \hline
sponge                                & 0.0567 & 0.0544 & 0.0584    & 0.0629 & 0.0580      \\
cup\_o\_noodles                       & 0.1049 & 0.1200 & 0.1029    & 0.0817 & 0.1509      \\
cloth\_hand\_cube                     & 0.0426 & 0.0767 & 0.0234    & 0.0288 & 0.0271      \\
cloth\_plastic\_bowl & 0.0619 & 0.1264 & 0.0544    & 0.0640 & 0.0604      \\
tupperware\_8oz                       & 0.1711 & 0.0982 & 0.0540    & 0.0665 & 0.0590      \\
timer\_in\_box                        & 0.0613 & 0.0935 & 0.0517    & 0.0650 & 0.0575      \\
red\_bottle\_square\_up               & 0.1886 & 0.1880 & 0.1092    & 0.1314 & 0.1009      \\
first\_aid                            & 0.1040 & 0.1169 & 0.0552    & 0.0480 & 0.0716      \\
foam\_red\_bowl                       & 0.0522 & 0.1108 & 0.0513    & 0.0428 & 0.0733      \\
green\_cup                            & 0.0624 & 0.0748 & 0.0503    & 0.0455 & 0.0612     
\end{tabular}
    \label{tab:Euclidean distance}
\caption{Mean euclidean distance error in m for each object}
\end{table}

\textbf{Action selection:} To ensure that we sample the actions from the distribution seen at train time, we initialize the mean and variance for the cross entropy method with the mean and variance of the actions in the train set. If the model being tested predicts that the action obtained from the optimizer will produce an end state that is more than 5cm (euclidean distance) away from the goal position retry the CEM method to obtain better actions.

\subsection{Latent Space Visualization}
\label{sec:ap_latent}
\begin{figure*}
    \includegraphics[width=0.5\textwidth]{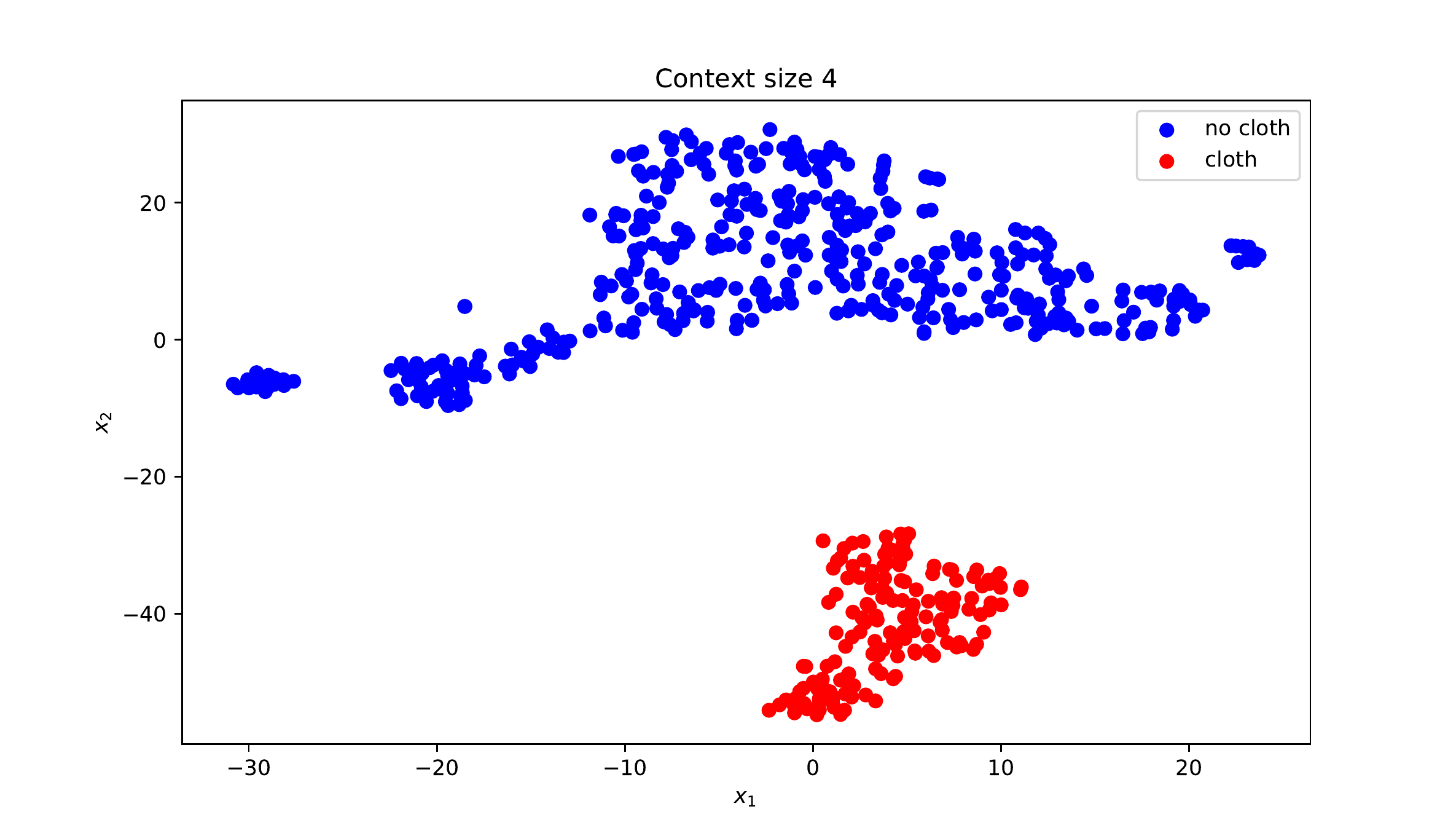}
    \includegraphics[width=0.5\textwidth]{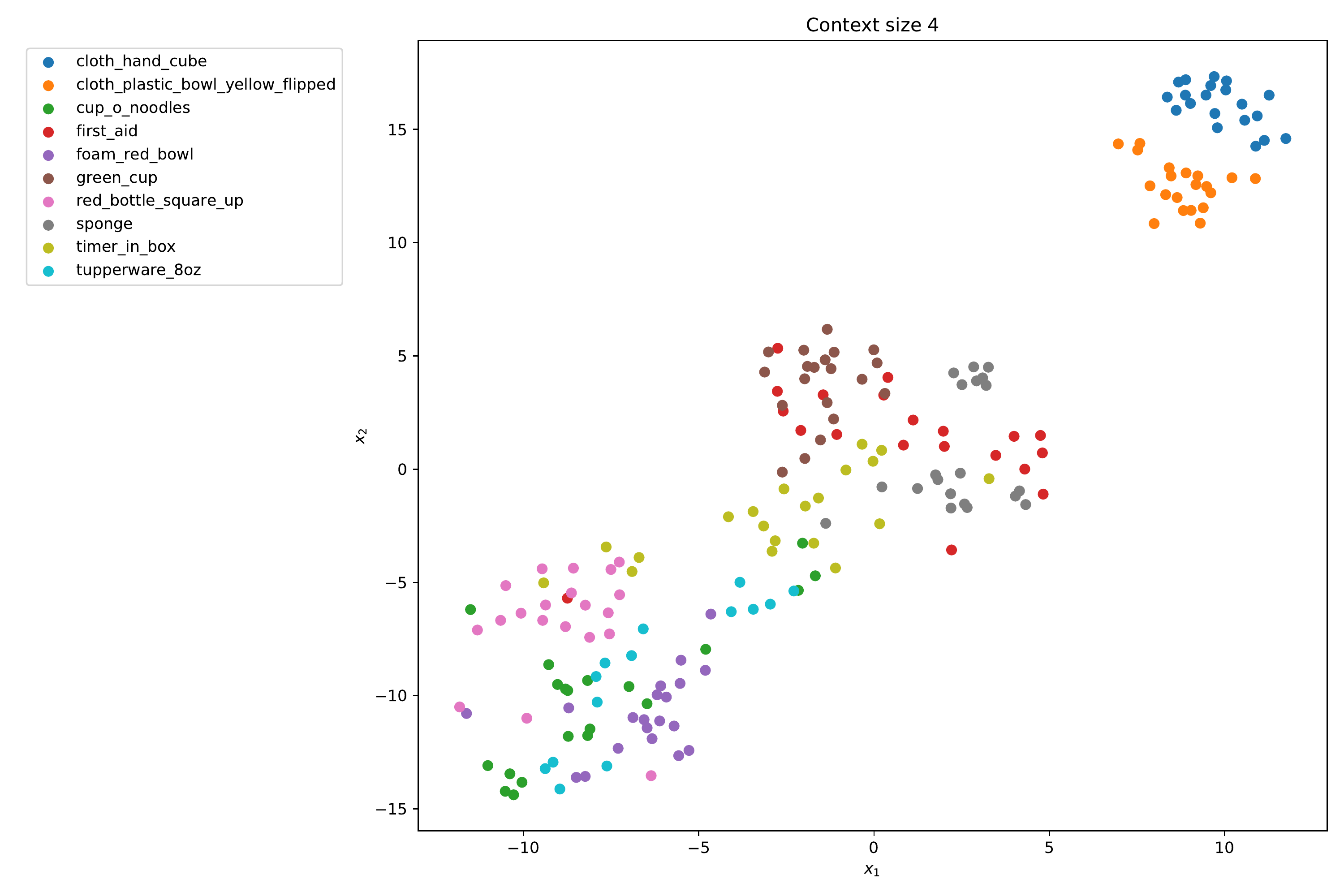}
    \caption{Left: t-SNE of latent space of train objects with objects slid on cloth in red. They are separable from those slid without cloth. Right: t-SNE of latent space of transformer model on the test objects.}
    \label{fig:tfm_latent_test_cloth}
\end{figure*}
\begin{figure*}
    \includegraphics[width=0.5\textwidth]{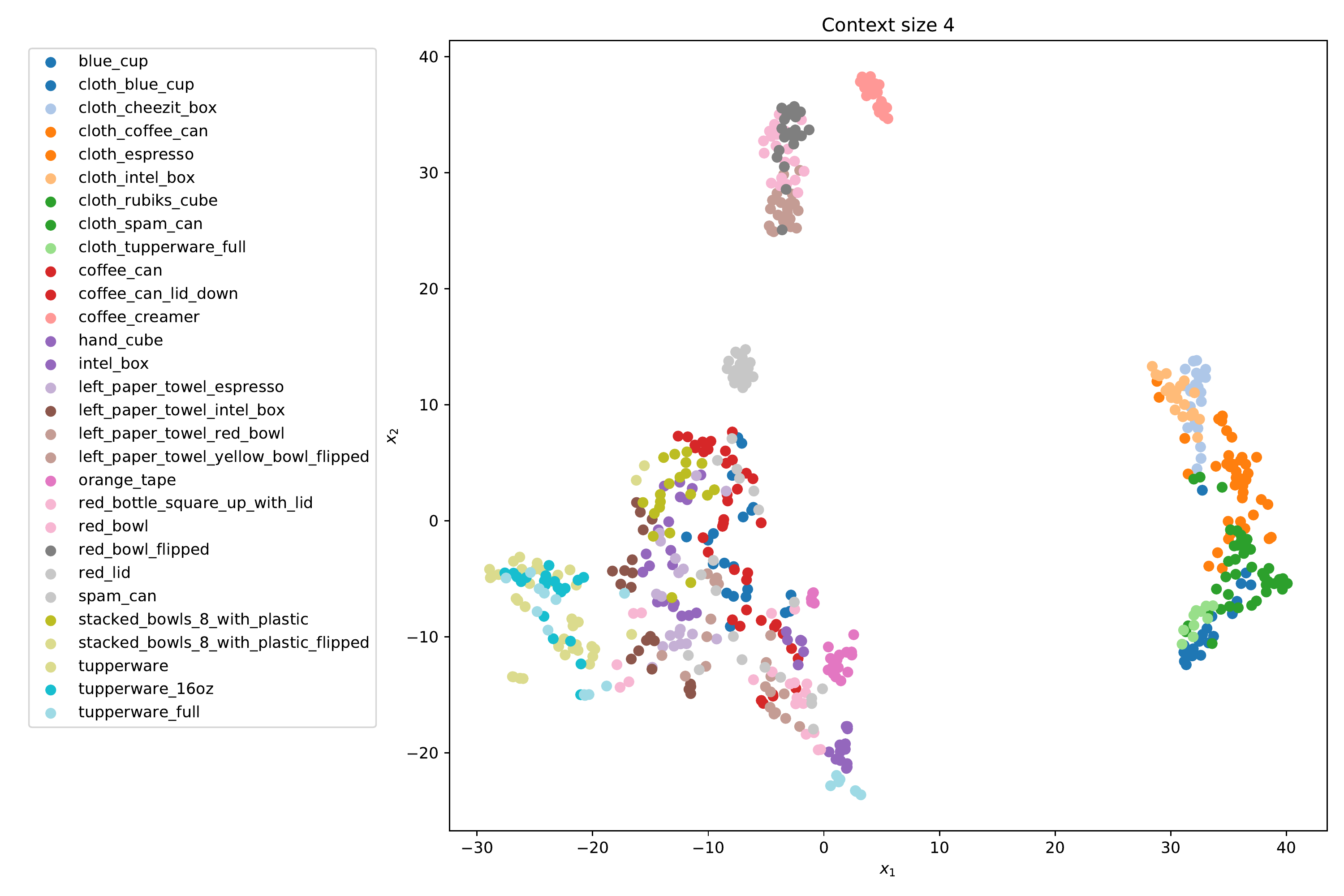}
    \includegraphics[width=0.5\textwidth]{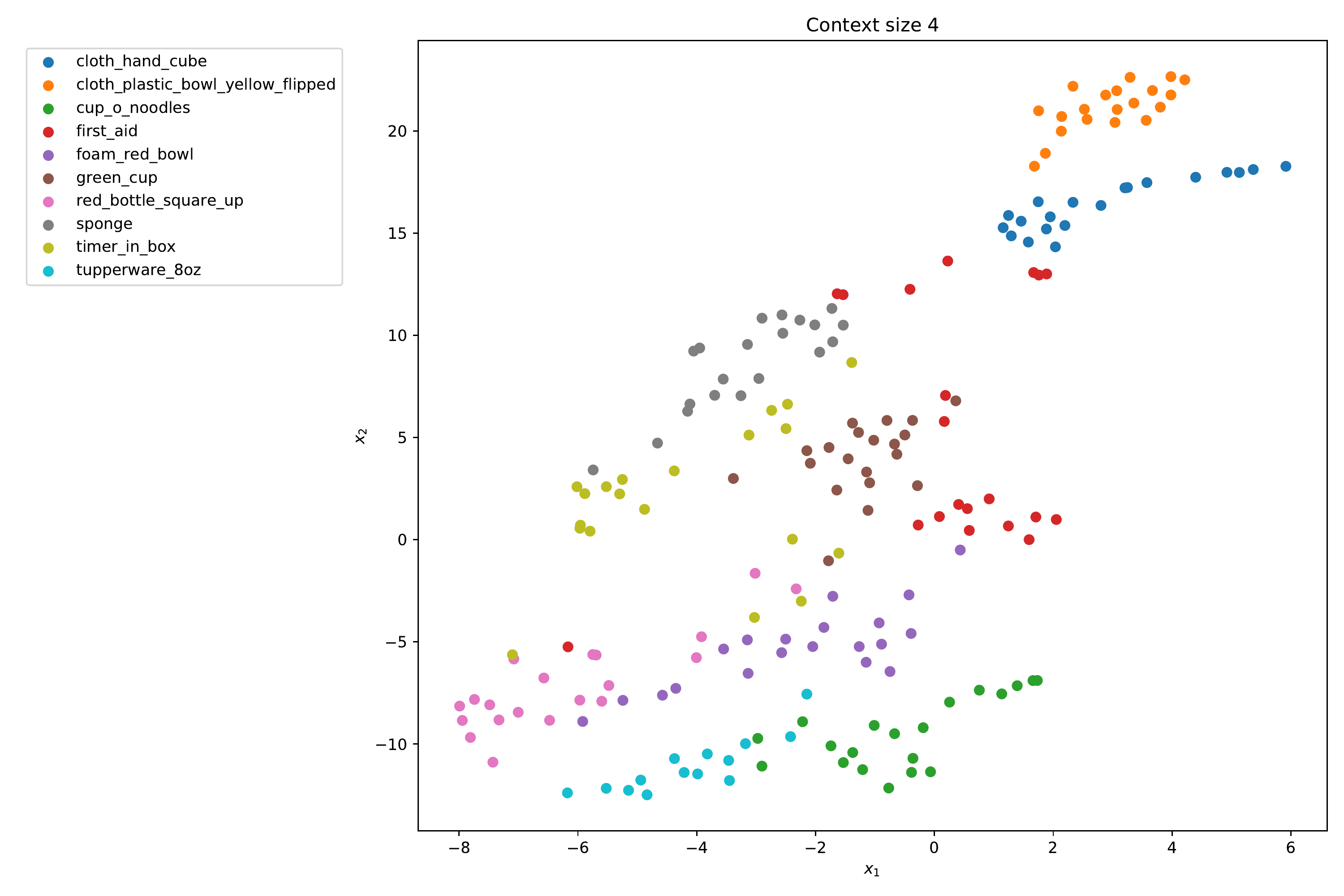}
    \caption{t-SNE of latent space of RNN model on the train and test objects.}
    \label{fig:rnn_latent}
\end{figure*}
\begin{figure*}
    \includegraphics[width=0.5\textwidth]{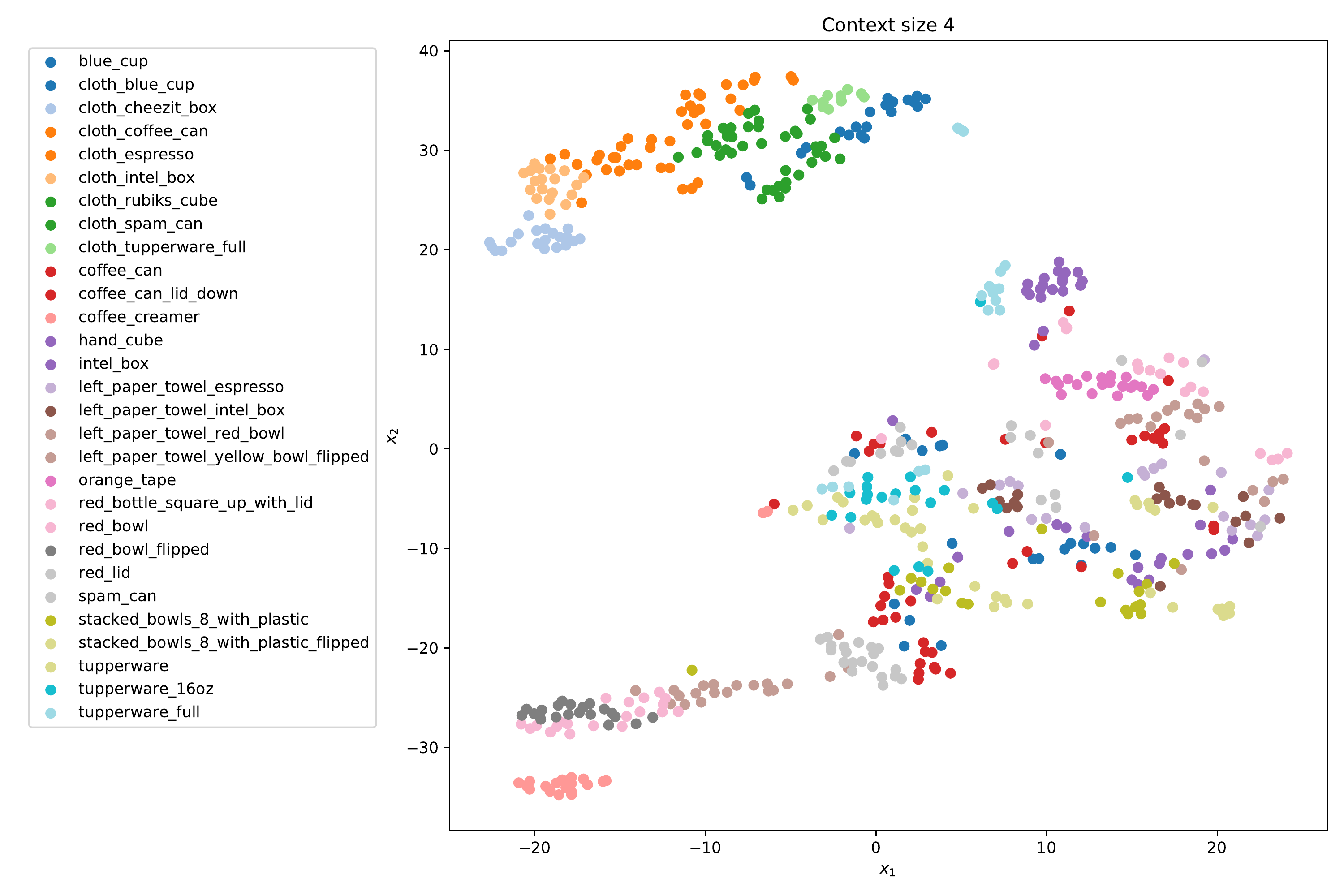}
    \includegraphics[width=0.5\textwidth]{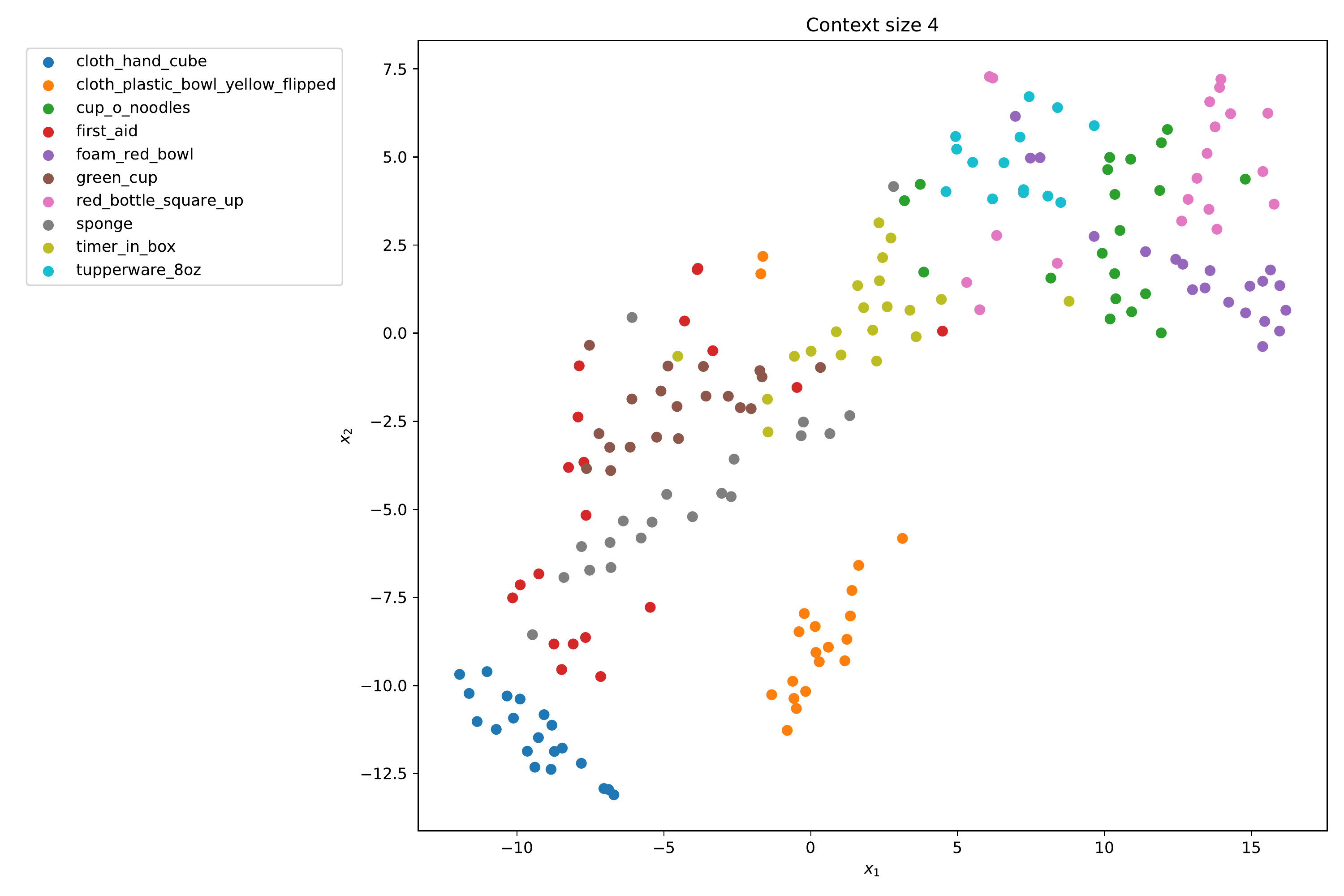}
    \caption{t-SNE of latent space of Continuous model on the train and test objects.}
    \label{fig:cont_latent}
\end{figure*}

\begin{figure*}
    \includegraphics[width=\textwidth]{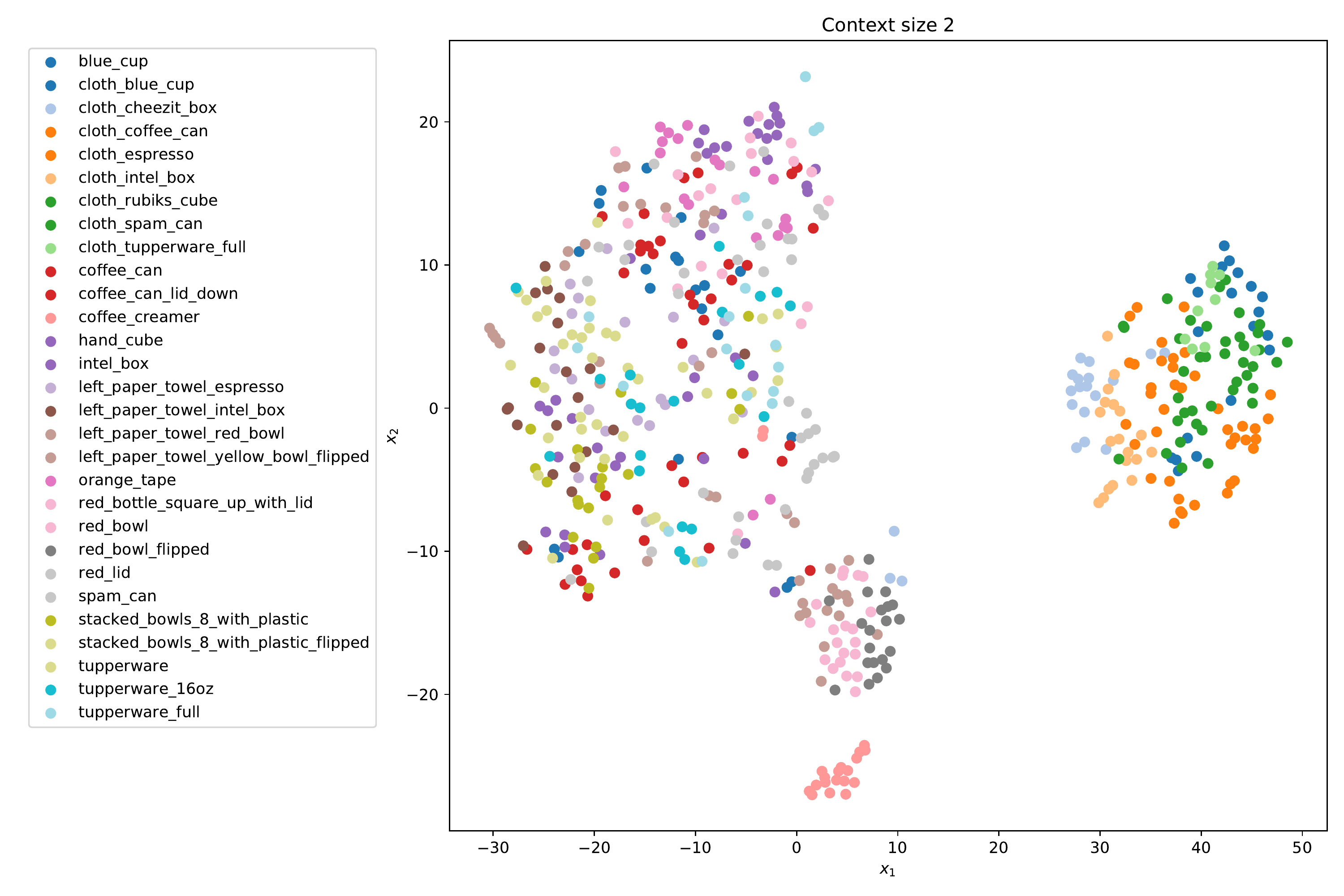}
    \includegraphics[width=\textwidth]{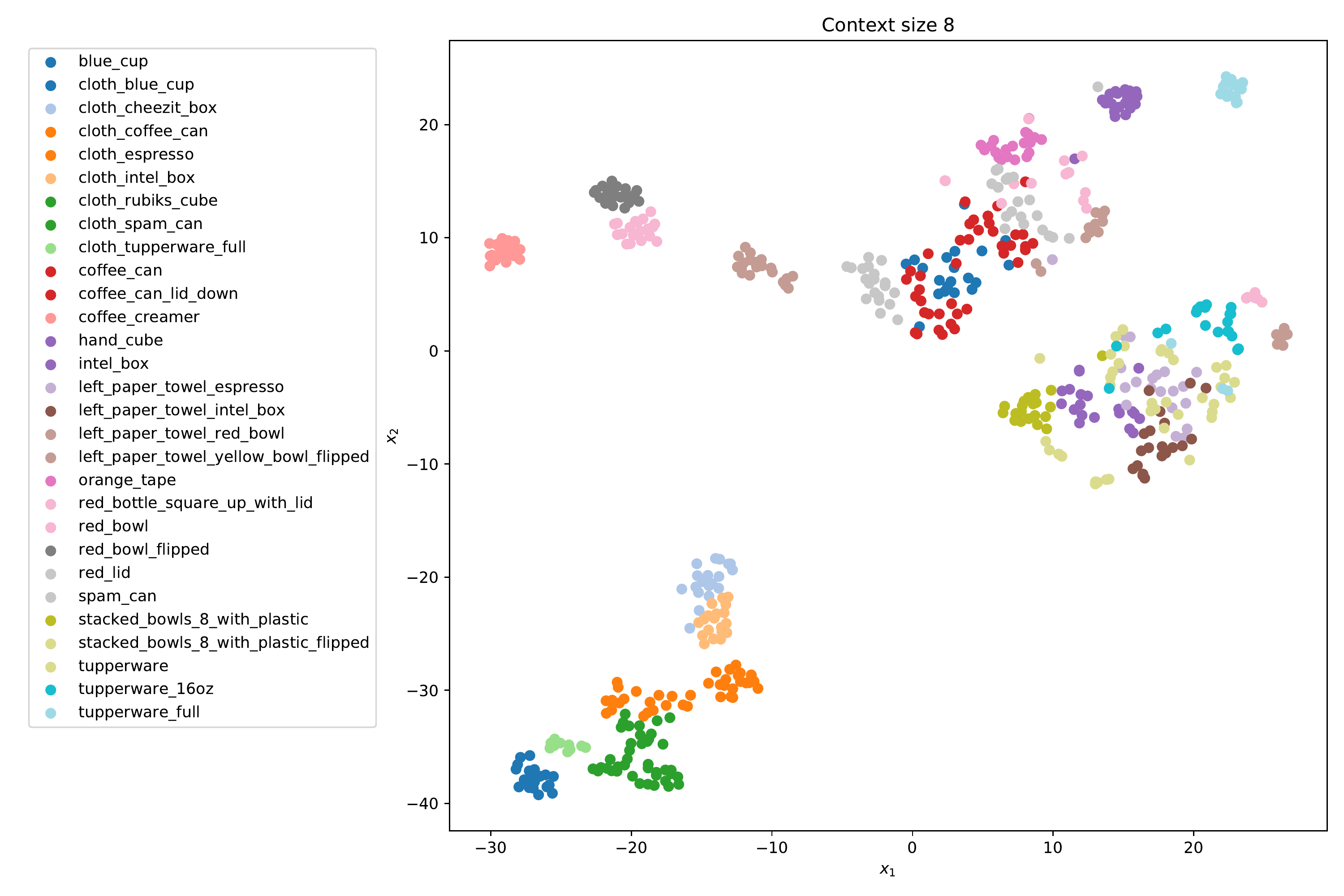}
    \caption{Latent space of transformer model on the train objects with 2 and 8 context points. Given more context points, the latent $z$'s generated are better separable from each other and more tightly clustered.}
    \label{fig:cont_latent}
\end{figure*}

\end{document}